\documentclass[twoside]{article}
\usepackage{booktabs}
\usepackage[pdftex]{graphicx}
\usepackage{amsmath}
\usepackage{amsthm}
\usepackage{amsfonts}
\usepackage{footnote}
\usepackage[accepted]{aistats2016}
\DeclareMathOperator*{\argmin}{arg\,min}
\begin{document}

% If your paper is accepted and the title of your paper is very long,
% the style will print as headings an error message. Use the following
% command to supply a shorter title of your paper so that it can be
% used as headings.
%
%\runningtitle{I use this title instead because the last one was very long}

% If your paper is accepted and the number of authors is large, the
% style will print as headings an error message. Use the following
% command to supply a shorter version of the authors names so that
% they can be used as headings (for example, use only the surnames)
%
%\runningauthor{Surname 1, Surname 2, Surname 3, ...., Surname n}

\twocolumn[

\aistatstitle{Adopting Robustness and Optimality in Fitting and Learning}

\aistatsauthor{Zhiguang Wnag, Tim Oates \And James Lo }
%\aistatsauthor{Author}

\aistatsaddress{ Department of CS and EE \\ University of Maryland, Baltimore County \And Department of Mathematics and Statistics \\ University of Maryland, Baltimore County } ]
%\aistatsaddress{University}]

\begin{abstract}
We generalized a modified exponentialized estimator by pushing the robust-optimal (RO) index $\lambda$ to $-\infty$ for achieving robustness to outliers by optimizing a quasi-Minimin function. The robustness is realized and controlled adaptively by the RO index without any predefined threshold. Optimality is guaranteed by expansion of the convexity region in the Hessian matrix to largely avoid local optima. Detailed quantitative analysis on both robustness and optimality are provided. The results of proposed experiments on fitting tasks for three noisy non-convex functions and the digits recognition task on the MNIST dataset consolidate the conclusions. 
\end{abstract}

\section{Introduction}
Function approximation has many applications in science and engineering, such as machine learning,  pattern recognition, signal processing and control theory. As universal approximators \cite{hornik1989multilayer}, neural networks (NNs) often deliver very good performance and have become the standard for several machine learning tasks given their recent success in various applications \cite{jaderberg2015spatial,lee2014deeply,szegedy2014going,Karpathy_2015_CVPR}.  

Error-free data are rarely provided in applications. Instead, data are usually contaminated by noise and outliers. Noise reflects inaccuracies in observations and the stochastic nature of the underlying process. NNs deal with such noise quite efficiently by optimizing the minimum squared error (MSE) or cross entropy (CE) between the observed and predicted values/labels with $\ell_1/\ell_2$ regularization, sparsity or adversary resistant learning \cite{sra2012optimization}. Recent developments in computer vision take advantage of manually added noise and distortion to further enhance the data \cite{krizhevsky2012imagenet,gan2015learning} or to capture reverse conditional probability within generative models \cite{bengio2013generalized,goodfellow2014generative}. Outliers can be arbitrary, unbounded, and not from any specific distribution, reflecting sudden abnormal changes or a mixture of rare but different phenomena. It has been noted that outliers in routine data are infrequent but heavily impact even high dimensional data \cite{aggarwal2001outlier,kriegel2008angle}. %This differentiates outliers with noise since learning with a bunch of noise sometimes help in learning a more robust model. 
When fitting a model or learning an objective function with rare but significant outliers whose distribution and impact are both unknown, they need to be identified and eliminated to get rid of their effect. Otherwise, the model overfits or underfits easily if training procedure is sufficient or not. For example, in the image/video classification task, some images or videos may be corrupted unexpectedly due to the error of sensors or severe occlusions of objects. Such outliers can skew parameter estimation severely and hence destroy the performance of the learned model. Alternatively, outliers are sometimes supposed to be examined closely, as they may be of interest themselves given the sample applications of intrusion or fraud detection. 

The maximal likelihood principle is not robust against outliers among different approximation and learning models like logistic regression (LR) and \cite{feng2014robust} generalized linear models \cite{kunsch1989conditionally}. High breakdown methods have been developed including least median of squares \cite{rousseeuw1984least}, least trimmed squares \cite{rousseeuw1984least} and most recently trimmed inner product methods for linear regression. Several works have investigated multiple approaches to robustify LR \cite{pregibon1982resistant,stefanski1986optimally,tibshirani2013robust}. The majority of them are M-estimator based: minimizing a complicated and more robust loss function than the standard loss function (negative log-likelihood). A robust backpropagation algorithm is proposed to optimize a robust estimator derived from the M-estimator to train NNs \cite{chen1994robust}. \cite{liano1996robust} used M-estimators to study the mechanism by which outliers affect the resulting NNs. The majority of the above work that is M-estimator based  needs to predefine a threshold for determining the degree of contaminated data to achieve robustness to outliers. Besides, few of them discuss optimality while achieving the robustness. Here, by optimality, we refer to its ability to find a nearly global or better local optimum while achieving robustness to outliers by getting rid of large deviation among the data. 

Our work is largely inspired by the exponentialized error criteria \cite{jacobson1973optimal,whittle1990risk} and its recent variations and modifies exponentialized error with two different training methods \cite{lo2010convexification,gui2014pairwise,wang2015adaptive}. Our main contribution is generalizing this estimator by pushing the robust-optimal (RO) index $\lambda$ to $-\infty$ to obtain the robustness to the largely deviated outliers while preserving the optimality by the expansion of the convexity regions in the Hessian matrix. As a general error estimator, we provide a quantitative analysis and validate its effectiveness on three function fitting and one visual recognition tasks.    

\section{Normalized Anomaly-Avering Estimator}
\newtheorem{mythm}{Theorem}
\newtheorem{mydef}{Definition}

Given standardized training samples $\{\boldsymbol{X},y\} = \{(\boldsymbol{x}_1, y_1), (\boldsymbol{x}_2, y_2), ..., (\boldsymbol{x}_m, y_m)\}$, $f(\boldsymbol{x}_i, \boldsymbol{W})$ is the learned model with parameters $W$. The loss function of mean squared loss (MSE) is defined as: 
\begin{eqnarray}
l_2(f(\boldsymbol{x}_i,\boldsymbol{W}), y_i) = \frac{1}{m}\sum_{i=1}^{m}(f(\boldsymbol{x}_i,\boldsymbol{W})-y_i)^2
\label{eqn:lp-error}
\nonumber
\end{eqnarray}

As a modified exponentialized error of MSE, the Anomaly-Averting Estimator (AAE) is defined as
\begin{eqnarray}
AAE = \frac{1}{m}\sum_{i=1}^{m}e^{\lambda (f(\boldsymbol{x}_i,\boldsymbol{W})-y_i)^2} & (\lambda<0) 
\label{eqn:lp-RAE}
\end{eqnarray}

The only difference between the AA estimator and RAE in \cite{lo2010convexification} is that $\lambda$ stays negative, which makes it serve as the robust-optimal (RO) index to control both the degree of optimality and robustness to outliers. Note that when $\lambda \to -\infty$, it is not bounded and suffers from less stability, exponential magnitude and arithmetic overflow when using gradient descent in implementations. We define the Normalized Anomaly-Averting Estimator (NAAE) as:
\begin{eqnarray} 
&& NAAE(f(\boldsymbol{x}_i,\boldsymbol{W}), y_i) \nonumber \\
&& = \frac{1}{\lambda}\log AAE(f(\boldsymbol{x}_i,\boldsymbol{W}), y_i) \nonumber \\
&& = \frac{1}{\lambda}\log \frac{1}{m}\sum_{i=1}^{m}e^{\lambda (f(\boldsymbol{x}_i,\boldsymbol{W})-y_i)^2}  (\lambda<0)
\label{eqn:NAAE}
\end{eqnarray}

\subsection{Robustness}
NAAE is a bounded estimator. The bounds exert not only the stable property as an estimator, but also an interesting dynamics to control the degree of robustness.   

\begin{mythm}%[Bounded]
	$NAAE(f(\boldsymbol{x}_i,\boldsymbol{W}), y_i)$ is bounded between the minimal and maximal squared errors.
\end{mythm}

\begin{proof}
	Let 
	\begin{eqnarray}
	\alpha_i=f(\boldsymbol{x}_i,\boldsymbol{W})-y_i \label{eqn:rawerror} \nonumber\\  \alpha_{min}^2 = \min \{\alpha_i^2, i = 1, ..., m\} \nonumber\\
	\alpha_{max}^2 = \max \{\alpha_i^2, i = 1, ..., m\} \nonumber\\
	\beta_{i} = e^{\lambda(\alpha_i^2-\alpha_{min}^2)} \nonumber
	\end{eqnarray}
	Then we can re-write Eqn. \ref{eqn:NAAE} as  
	\begin{eqnarray}
	&& NAAE \nonumber\\
	&& =\frac{1}{\lambda} \log \frac{1}{m} e^{\lambda\alpha_{min}^2}\sum_{i=1}^{m} \beta_i \nonumber \\
	&& =  -\frac{1}{\lambda} \log m + \alpha_{min}^2 + \frac{1}{\lambda}\log\sum_{i=1}^{m}\beta_i \label{eqn:demolition}
%	&\leq& -\frac{1}{\lambda} \log m + ||\alpha_{min}||^2 + \frac{1}{\lambda}\log m \cdot e^{\lambda||\alpha_{min}||^2} \nonumber \\
%	&=& 2||\alpha_{min}||^2
	\end{eqnarray}
	let $G=\alpha_{max}^2 - \alpha_{min}^2$ to be the maximal margin of the sample squared errors, considering $\lambda<0$, $e^{\lambda G} \leq \beta_i \leq 1$, so 
	\begin{eqnarray}
	\frac{1}{\lambda}\log m \leq \frac{1}{\lambda}\log\sum_{i=1}^{m}\beta_i \leq \frac{1}{\lambda}\log m + G \nonumber
	\end{eqnarray}
	We can derive the bounds of NAAE as 
	\begin{equation}
	\alpha_{min}^2 \leq NAAE \leq \alpha_{max}^2	\nonumber
	\end{equation}
\end{proof}

The above conclusions indicate that squared errors regulate the bounds of NAAE. The next theorem shows that $\lambda$ controls NAAE to perform interactively with MSE and its relationship to the minimin operator.      

\begin{mythm}
 when $\lambda \to -\infty$, $NAAE(f(\boldsymbol{x}_i,\boldsymbol{W}), y_i)$ approaches to an minimin operator; if $\lambda \to 0$, $NAAE(f(\boldsymbol{x}_i,\boldsymbol{W}), y_i) \to$ MSE.
\end{mythm}

\begin{proof}
Given $\lambda<0$, we consider the objective function 
\begin{eqnarray}
W = \argmin_{\boldsymbol{W}} NAAE(f(\boldsymbol{x}_i,\boldsymbol{W}), y_i)  \nonumber
\end{eqnarray}
Considering equation \ref{eqn:demolition}, it is easy to see that when $\lambda \to -\infty$, the objective function becomes
\begin{eqnarray}
W = \argmin_{\boldsymbol{W}} \alpha_{min}^2  \nonumber
\end{eqnarray}
So a large $\lambda$ controls the NAAE to approach a minimin operator. Next we consider the situation when $\lambda \to 0$ \footnote{Consider the rules $e^x = 1 + x + \frac{x^2}{2}+ \cdots (x \to 0)$ and $\log x = x - \frac{x^2}{2}+ \cdots (x \to 0)$}.  
\begin{eqnarray}
&& NAAE(f(\boldsymbol{x}_i,\boldsymbol{W}), y_i) \nonumber\\
&& = \frac{1}{\lambda}\log \frac{1}{m}\sum_{i=1}^{m}e^{\lambda(f(\boldsymbol{x}_i,\boldsymbol{W})-y_i)^2} \nonumber \\
&& = \frac{1}{\lambda}\log (1+ \frac{1}{m}\sum_{i=1}^{m}\lambda(f(\boldsymbol{x}_i,\boldsymbol{W})-y_i)^2 + O(\lambda^{2})) \nonumber \\
&& = \frac{1}{m}\sum_{i=1}^{m}(f(\boldsymbol{x}_i,\boldsymbol{W})-y_i)^2 \label{eqn:MSEalike}
\end{eqnarray}
\end{proof}

More rigid proofs that can be generalized to $L_p$-norm error are given in \cite{lo2010convexification}. 

Equations \ref{eqn:demolition} and \ref{eqn:MSEalike} explain how $\lambda$ controls the robustness level. When $\lambda \to 0$, NAAE approaches MSE with a breakdown point of 0\%,  meaning that a single observation can change it arbitrarily. Further, it is highly influenced by outliers. When $\lambda \to -\infty$, NAAE performs like a minimin operator to address only the minimal error. Outliers can be eliminated if they largely deviate from the objective space, or say manifold, so that NAAE concentrates on the smaller errors and ignores the impact of those large errors. But it is still vulnerable to noise and outliers that are much closer to the objective manifold than the majority of the clean data. 

When $\lambda$ changes between 0 and $-\infty$, the exponential sum of the squared error offsets the minimin operator to further address the situations when the outliers are close to the objective manifold. By adjusting $\lambda$, NAAE is able to handle both large deviations and false positive samples, which helps to achieve robustness to different type of outliers.

\subsection{Optimality}      

To prove the optimality or NAAE, we start with AAE and generalize our conclusion to NAAE. For clarity, we write the matrix derivatives in functional form. In our proof we use the quadratic form. The Jacobian matrix of Eqn. \ref{eqn:lp-RAE} is
\begin{eqnarray}
J(\boldsymbol{W}) =& \frac{2\lambda}{m} \sum_{i=1}^{m}e^{\lambda(f(\boldsymbol{x}_i,\boldsymbol{W})-y_i)^2} \nonumber \\ 
\times& (f(\boldsymbol{x}_i,\boldsymbol{W})-y_i)\frac{\partial f(\boldsymbol{x}_i,\boldsymbol{W})}{\partial \boldsymbol{W}} & (\lambda<0)
\label{eqn:Jacobian}
\end{eqnarray}

The Hessian matrix of Eqn. \ref{eqn:lp-RAE} is
\begin{eqnarray}
H(\boldsymbol{W}) =&  \frac{2}{m}\sum_{i=1}^{m}e^{\lambda(f(\boldsymbol{x}_i,\boldsymbol{W})-y_i)^2}
\{\lambda \frac{\partial f(\boldsymbol{x}_i,\boldsymbol{W})}{\partial \boldsymbol{W}}^2 \label{A} \\ 
+& 2\lambda^{2}(y_i-f(\boldsymbol{x}_i,\boldsymbol{W}))^{2}\frac{\partial f(\boldsymbol{x}_i,\boldsymbol{W})}{\partial \boldsymbol{W}}^2 \label{B} \\ 
+& \lambda(y_i-f(\boldsymbol{x}_i,\boldsymbol{W}))\frac{\partial f^2(\boldsymbol{x}_i,\boldsymbol{W})}{\partial \boldsymbol{W}^2} \label{C}\}
\end{eqnarray}

We assume the training sample is standardized and the error space is a unit sphere $|\mathcal{R}^n \leq 1|$ (that is, $|y_i-f(\boldsymbol{x}_i,\boldsymbol{W}|<1$). Because AAE has the sum-exponential form, its Hessian matrix is tuned exactly by the RO index $\lambda$. The following theorem indicates the relation between the convexity index and its convexity region.
\begin{mythm}%[Convexity]
	Given the Anomaly-Averting Error criterion $AAE$, which is twice continuous differentiable. $J(\boldsymbol{W})$ and $H(\boldsymbol{W})$ are the corresponding Jacobian and Hessian matrix. As $\lambda \to \pm\infty$, the convexity region monotonically expands to the entire parameter space except for the subregion $S:=\{W \in \mathcal{R}^n | rank(H(\boldsymbol{W}))<n, H(\boldsymbol{W}<0)\}$. 
	\label{thm:RAE} 
\end{mythm}

\begin{proof}
Both  $\frac{\partial f(\boldsymbol{x}_i,\boldsymbol{W})^2}{\partial \boldsymbol{W}}$ and $\lambda^{2}(y_i-f(\boldsymbol{x}_i,\boldsymbol{W}))^{2}$ are positive semi-definite, matrix \ref{B} is semi-positive definite, but Eqn. \ref{A} and \ref{C} may be indefinite. Let $\alpha_i(\boldsymbol{W})=f(\boldsymbol{x}_i,\boldsymbol{W})-y_i$, We rewrite Eqn. \ref{B} in quadratic form:  
\begin{eqnarray}
& = & p\lambda^{2}\alpha_i(\boldsymbol{W})^T\frac{\partial f(\boldsymbol{x}_i,\boldsymbol{W})^T}{\partial \boldsymbol{W}}\cdot\frac{\partial f(\boldsymbol{x}_i,\boldsymbol{W})}{\partial \boldsymbol{W}}\alpha_i(\boldsymbol{W}) \nonumber \\
& = & p\lambda^{2}\alpha_i(\boldsymbol{W})^T Q \Lambda Q^T\alpha_i(\boldsymbol{W}) \nonumber \\
& = & p\lambda^{2} S(\boldsymbol{W})^T \Lambda S(\boldsymbol{W}) \label{eqn:orthofinal}
\end{eqnarray}  

$\Lambda = diag[\Lambda_1, \Lambda_2, ..., \Lambda_m]$. \ref{A} is positive semi-definite. 

If $S(\boldsymbol{W})$ is a full-rank matrix, then Eqn. \ref{eqn:orthofinal} is positive definite. When $\lambda \to \pm\infty$, the eigenvalues $\Lambda$ becomes dominant in the leading principal minors (as well as eigenvalues) of the Hessian matrix to make $H(\boldsymbol{W})$ monotonically increasing with $\lambda$. Assume $P_\lambda:=\{W \in \mathcal{R}^n|H_(\boldsymbol{W}>0)\}$, we have $P_{\lambda_1} \in P_{\lambda_2}$ when $\lambda_1 < \lambda_2$. Considering Eqn. \ref{A}, \ref{B} and \ref{C} are all bounded, $\exists$ $\psi(W)$, s.t. $H(\boldsymbol{W}) > 0$ when $|\lambda|> \psi(W)$. 

When $S(\boldsymbol{W})$ is not a full-rank matrix, the determinant of all its $n \choose k$ submatrices is 0. Thus, in the subregion $S:=\{W \in \mathcal{R}^n | rank(H(\boldsymbol{W}))<n, H(\boldsymbol{W}<0)\}$, there is no parameters satisfied the convexity conditions ($\bigcup P_\lambda$).
\end{proof}

Theorem \ref{thm:RAE} states that when the RO index $\lambda$ decrease to infinity, the convexity region in the parameter space of AAE expands monotonically to the entire space except the intersection of a finite number of lower dimensional sets. The number of sets increases rapidly as the number $m$ of training samples increases. Roughly speaking, large $|\lambda|$ and $m$ cause the size of the convexity region to grow larger in the error space of AAE.

When $\lambda \to -\infty$, the error space can be perfectly stretched to be strictly convex, thus avoiding the local optimum to find a global optimum. The following theorem states the quasi-convexity of NAAE. 

\begin{mythm}%[Quasi-convexity]
	Given a parameter space $\{W \in \mathcal{R}^n \}$, Assume $\exists$ $\psi(W)$, s.t. $H(\boldsymbol{W}) > 0$ when $|\lambda|> \psi(W)$ to guarantee the convexity of $AAE(f(\boldsymbol{x}_i,\boldsymbol{W}), y_i)$. Then, $NAAE(f(\boldsymbol{x}_i,\boldsymbol{W}), y_i)$ is  \textbf{quasi-convex} and share the same local and global optima with $AAE(f(\boldsymbol{x}_i,\boldsymbol{W}), y_i)$.
	\label{thm:quasi}
\end{mythm}
\begin{proof}
	If $AAE(f(\boldsymbol{x}_i,\boldsymbol{W}), y_i)$ is convex, it is quasi-convex. The $\log$ function is monotonically increasing, so the composition  $\log AAE(f(\boldsymbol{x}_i,\boldsymbol{W}), y_i)$ is quasi-convex. \footnote{Because the function $f$ defined by $f(x) = g(U(x))$ is quasi-convex if the function $U$ is quasiconvex and the function $g$ is increasing.}  
	
	$\log$ is a strictly monotone function and $NAAE(f(\boldsymbol{x}_i,\boldsymbol{W}), y_i)$ is quasi-convex, so it shares the same local and global optima with $AAE(f(\boldsymbol{x}_i,\boldsymbol{W}), y_i)$.
\end{proof}

The above theorem states that the convexity region of NAAE is consistent with AAE. To interpret this statement in another perspective, the $\log$ function is a strictly monotone function. Even  if AAE is not strictly convex, NAAE still shares the same local and global optima with RAE. If we define the mapping function $f:AAE \to NAAE$, it is easy to see that $f$ is  bijective and continuous. Its inverse map $f^{-1}$ is also continuous, so that $f$ is an open mapping. Thus, it is easy to prove that the mapping function $f$ is a homeomorphism to preserve all the topological properties of the given space.

The above theorems state the consistent relations among NAAE, AAE and MSE. It is proven that the greater the RO index $|\lambda|$, the larger the convex region is. Intuitively, increasing $|\lambda|$ creates tunnels for a local-search minimization procedure to travel through to a good local optimum.  While NAAE preserves the robustness to MSE, theorem \ref{thm:lowerbound} provides insights into when NAAE is optimal than MSE.      

\begin{mythm}%[Lower-bound]
	Given training samples $\{\boldsymbol{X},y\} = \{(\boldsymbol{x}_1, y_1), (\boldsymbol{x}_2, y_2), ..., (\boldsymbol{x}_m, y_m)\}$ and the model $f(\boldsymbol{x}_i, \boldsymbol{W})$ with parameters $W$. Define the deviation $\alpha_i=f(\boldsymbol{x}_i,\boldsymbol{W})-y_i$, when $\alpha_i \to 1$ and $\lambda \leq -1$, both $AAE(f(\boldsymbol{x}_i,\boldsymbol{W}), y_i)$ and  $NAAE(f(\boldsymbol{x}_i,\boldsymbol{W}), y_i)$ always have a larger convexity region to commit a higher chance for the better local optima than MSE.
	\label{thm:lowerbound}
\end{mythm}

\begin{proof}
	Let $h(\boldsymbol{W})$ denotes the Hessian matrix of MSE (Eqn. \ref{eqn:lp-error}),
	\begin{eqnarray}
	h(\boldsymbol{W}) &=&\frac{2}{m}\sum_{i=1}^{m}\{\alpha_i(\boldsymbol{W})^{2} \frac{\partial f(\boldsymbol{x}_i,\boldsymbol{W})^2}{\partial \boldsymbol{W}}  \nonumber \\ &+& \alpha_i(\boldsymbol{W})\frac{\partial f^2(\boldsymbol{x}_i,\boldsymbol{W})}{\partial \boldsymbol{W}^2}\}
	\end{eqnarray}
	
	Since $\lambda \leq -1$, let $diag_{eig}$ denote the diagonal matrix of the eigenvalues from SVD decomposition. $\succeq$ here means 'element-wise greater'. When $A \succeq B$, each element in A is greater than B. Then we have
	\begin{eqnarray}
	&& diag_{eig}[H(\boldsymbol{W})]  \nonumber\\ 
	&& \succeq \frac{2}{m}\{(2\lambda^2\alpha_i^2+\lambda)\frac{\partial f(\boldsymbol{x}_i,\boldsymbol{W})^2}{\partial \boldsymbol{W}} - \lambda\alpha_i \frac{f^2(\boldsymbol{x}_i,\boldsymbol{W})}{\partial \boldsymbol{W}^2} \nonumber
		\} \nonumber \\
	&& \succeq \alpha_i^2\frac{f(\boldsymbol{x}_i,\boldsymbol{W})^2}{\partial \boldsymbol{W}} + \alpha_i\frac{f^2(\boldsymbol{x}_i,\boldsymbol{W})}{\partial \boldsymbol{W}^2} \nonumber \nonumber\\
	&& \succeq diag_{eig}[h(\boldsymbol{W})] \nonumber
	\end{eqnarray}	
	
Briefly, when the standard deviation is large (approaches 1) and $\lambda \leq -1$, $AAE(f(\boldsymbol{x}_i,\boldsymbol{W}), y_i)$ always has larger convexity regions than MSE to better enable escape of local minima. Because $NAAE(f(\boldsymbol{x}_i,\boldsymbol{W}), y_i)$ is quasi-convex, sharing the same local and global optimum with $AAE(f(\boldsymbol{x}_i,\boldsymbol{W}), y_i)$, the above conclusions are still valid.  
\end{proof}

The expansion of the convexity region in the Hessian matrix enables NAAE to find a better local optima than MSE. This property guarantees higher probability to escape poor local optima. In the worst case, NAAE will perform as good as MSE if the convexity region shrinks as the RO index $\lambda$ increase to approach 0 or the local search deviates from the "tunnel" of convex regions.   

\subsection{Control Robustness and Optimality}
NAAE performs robustly while preserving the optimality by the expansion of its convex region in its Hessian matrix, the RO index $\lambda$  controls both the robustness and optimality of the estimator simultaneously.  When $\lambda$ is a large negative number, NAAE works like a minimin operator to avoid large deviations incurred by the outliers. Meanwhile, its Hessian matrix has a larger convex region to facilitate seeking a better optimum. If $\lambda$ approaches 0, a compensation from the exponential sum of the squared errors offsets the impact of the minimin operator to further focus on the larger overview. This will help to address the small deviation from other noise. NAAE performs more like  MSE to grant a smooth and stable optimization procedure.      

\begin{table*}[t]
	\centering
	\caption{Training and test MSE of approximation on the three sample functions. For NAAE, $\lambda$ is updated by the fixed-$\lambda$ and adaptive learning strategies. The average performance over 10 runs are reported.}
	\begin{tabular}{lllllll}
		\toprule
		& \multicolumn{2}{c}{MSE} & \multicolumn{2}{c}{NAAE-fixed-$\lambda$} & \multicolumn{2}{c}{NAAE-adaptive learning} \\
		\midrule
		& Training  & Test  & Training & Test  & Training  & Test  \\
		$F_1(x)$ & 0.02  & 0.022 & 0.021 & 0.019 & 0.021 & \textbf{0.018} \\
		$F_2(x)$ & 0.013 & 0.029 & 0.005 & \textbf{0.005} & 0.005 & \textbf{0.005} \\
		$F_3(x)$ & 0.012 & 0.037 & 0.045 & \textbf{0.01} & 0.045 & \textbf{0.01} \\
		\bottomrule
	\end{tabular}%
	\label{tab:funcapproxstats}%
\end{table*}%

\section{Update Strategies}
As $\lambda$ controls the degree of both robustness and optimality, the initialization and update strategy impacts the learning performance. Generally, large $\lambda$ is preferable at the start. Different update strategies lead to three major learning methods.

\subsection{Fixed-$\lambda$}
Using a fixed large $\lambda$ to optimize the exponential estimator is reported in \cite{lo2012overcoming}. For NAAE, a fixed $\lambda$ consistently controls the estimator at a specific level of the offset towards the minimin operator. The weights are updated using gradient descent. It is simple and work well to approximate the lower dimensional functions with outliers. When attacking the learning tasks in higher dimensions such as visual recognition, the large plateau and unstable learning procedure can lead to a failure in training \cite{wang2015adaptive,gui2014pairwise}.   

\subsection{Gradual Deconvexfication}
\cite{lo2013overcoming} proposed the Gradual Deconvexification (GDC) approach to alleviate the difficulty in finding a good value of $\lambda$ for training. GDC starts with a very large $\lambda$ while recording the values of the objective function before and after a pre-selected number of training epochs. If the performance converges, GDC flags the current training condition as stagnant and performs a deconvexification by reducing the current $\lambda$ with a percentage $r$. After that, the training continues and repeats the deconvexification if necessary until a satisfied training error is achieved. If $|\lambda|$ keeps coming down to below 1, the training will set $\lambda = 0$, which actually converts to the training with MSE. All other weights are updated using gradient descent.

GDC works in a similar way as the decreasing-learning-rate approach. Training with GDC avoids fixing the initial $\lambda$ during training, as the deconvexification gradually decreases $\lambda$ to avoid bad optima and vulnerable extreme estimator and eventually obtain a reasonable optimum while achieving robustness. As reported in \cite{gui2014pairwise}, GDC is much slower than using MSE alone. The update strategy relies on the per-defined convergence thresholds and dropping rate. 

\subsection{Adaptive Training}
\cite{wang2015adaptive} proposed a novel learning method to training with RO index $\lambda$, called the Adaptive Normalized Risk-Avering Training approach (adaptive training). Instead of manually tuning $\lambda$ like GDC , they learn $\lambda$ adaptively through error backpropagation by considering $\lambda$ as a parameter instead of a hyperparameter. The learning procedure is standard batch gradient descent.

\begin{eqnarray}
\frac{dl(W, \lambda)}{d\lambda} 
&\approx&  \frac{q}{\lambda}(L_P\text{-norm error}-E)
\label{eqn:nglambda}
\end{eqnarray} 

As shown in the above equation,  the gradient on $\lambda$ is approximately the difference between the exponentialized estimator $E$ (e.g., NAAE in this paper) and the standard $L_p$-norm error. Considering $\lambda<0$ as for NAAE. When $E$ is larger, $\lambda$ decreases to enlarge the convexity region, facilitating the search in the error space for better optima. When $E$ is smaller, the learned parameters are seemingly going through the optimal "tunnel" for better optima. $\lambda$ then increases and helps the weights not deviate far from the manifold of the standard $L_p$-norm error to make the error space stable without large plateaus. We note that adaptive training also adjusts $\lambda$ to control the robustness level in a similarly 'smart' way. If $E$ is larger than the $L_p$-norm error, $\lambda$ decreases to further ignore more samples with large deviations to let the learning function gain a broader overview rather than focusing on the outliers. When $E$ is small, outliers with large deviation are almost eliminated. $\lambda$ increases to approach 0, making the learning procedure more accurate and stable by considering the effect of more training samples rather than the extreme value. The adaptive training approach has more flexibility. It keeps searching the error space near the manifold of the $L_p$-norm error to find better optima in a way of competing with, and at the same time relying on, the standard $L_p$-norm error space.   

We consider the fixed $\lambda$ approach and adaptive training as the update strategies for the RO index $\lambda$ in our experiments. For Adaptive training, the final loss function is
\begin{eqnarray}
l(W,\lambda) = \frac{1}{\lambda}\log \frac{1}{m}\sum_{i=1}^{m}e^{\lambda(f(\boldsymbol{x}_i,\boldsymbol{W})-y_i)^2} + a |\lambda|^{-1}
\label{eqn:NAAEloss}
\end{eqnarray} 

The penalty weight $a$ controls the convergence speed by penalizing small $|\lambda|$. Smaller $a$ emphasizes tuning $\lambda$ to allow faster convergence speed between NAAE and MSE. Larger $a$ forces larger $|\lambda|$ for a better chance to find a better optimum while skipping the outliers, but runs the risk of plateaus and deviating far from the stable error space. 

\section{Experiments and Analysis}

\begin{figure*}[t]
	\vspace{.3in}
	\centering
	\includegraphics[width=7 in]{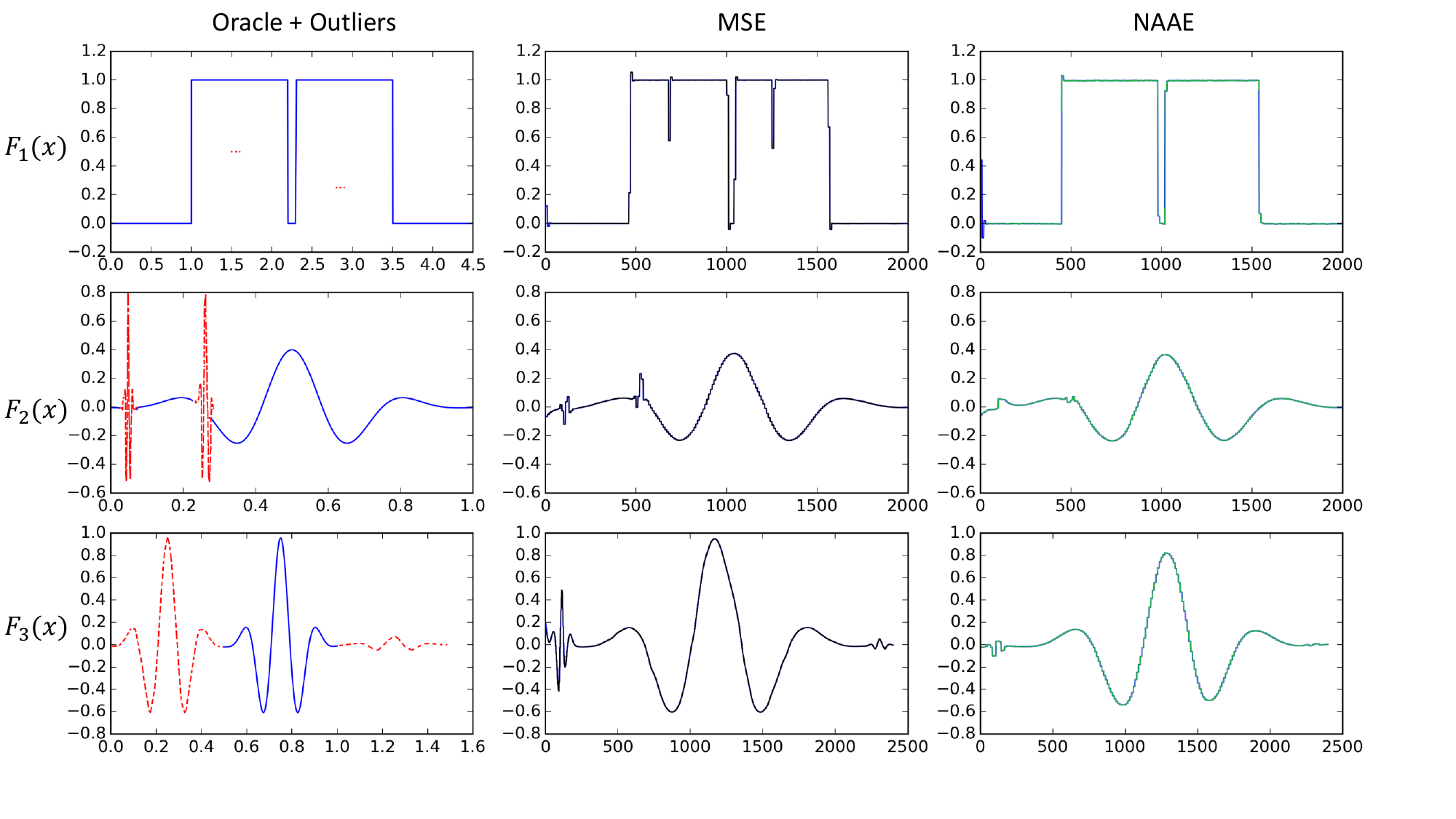}
	\caption{Function approximation using MSE and NAAE on the training set. Three columns are (a). oracle function with outliers (L), (b). approximation results using MSE (M) and (c). NAAE (R). In column (a), the blue lines plot the oracle function, the red dots plot the outliers.} 
	\label{fig:funcapprox}
	\vspace{.3in}
\end{figure*}
As an error estimator, we test NAAE on the function approximation and digits recognition tasks with two different neural networks architectures, multiple layer perceptrons (MLPs) and convolutional networks (ConvNets), respectively. We limit all the experiments in very simple settings, using gradient descent with batch gradient descent and weight decay or dropout \cite{srivastava2014dropout}. 

\subsection{Function Approximation}
For function approximation tasks, we design three typical non-convex functions with different types of outliers to test the robustness and optimality achieved by NAAE.

1). The notch function is defined by
\begin{eqnarray}
f_1(x) =
\begin{cases}
0       & \quad \mathrm{if} \quad x \in [0, 1.0] \bigcup [2.2, 2.3] \bigcup [3.5, 4.5] \\
1  & \quad \mathrm{otherwise} \\
\end{cases} \nonumber
\label{eqn:notchfunction}
\end{eqnarray}

The outliers in the middle range are generated by the function
\begin{eqnarray}
f_1^*(x) =
\begin{cases}
0.25      & \quad \mathrm{if} \quad x \in [2.8, 2.81] \\
0.5  & \quad \mathrm{if} \quad x \in [1.5, 1.51] \\
\end{cases} \nonumber
\label{eqn:notchfunctionoutlier}
\end{eqnarray}

where $ x \in X = [0,4.5]$. $x_i$ are obtained by random sampling 2000 non-repeatable numbers from $X$ with a uniform distribution, and the corresponding output values $y_k$ are computed by $f(x)$ and $f^*(x)$. The overall function is $F_1(x) = f_1(x)+f_1^*(x)$. The training data with 2000 $(x_i, y_i)$ pairs is chosen to perform the notch function approximation $f(x)$ with the corruption of the outliers $f^*(x)$. MLPs with 1:16:1 architectures are initiated to all training sessions.  

2). The smooth function is defined by
\begin{eqnarray}
f_2(x) = g(x, \frac{1}{6}, \frac{1}{2}, \frac{1}{6}) \nonumber
\end{eqnarray}

The outliers with very large derivations  given by 
\begin{eqnarray}
f_2^*(x) = g(x, \frac{1}{64}, \frac{1}{4}, \frac{1}{128}) + g(x, \frac{1}{64}, \frac{1}{20}, \frac{1}{128}) \nonumber
\end{eqnarray}
Where $x \in X = [0,1]$, $g$ is defined as 
\begin{eqnarray}
g(x) = \frac{\alpha}{\sqrt{2\pi}\tau} \cos(\frac{(x-\mu)\pi}{\tau})e^{-\frac{(x-\mu)^2}{2\tau^2}}
\label{eqn:g}
\end{eqnarray}

The overall function is $F_2(x) = f_2(x)+f_2^*(x)$. The input values $x_i$ are selected by sampling 2000 numbers from a uniformly distributed grid on $X$. The corresponding output values $y_i$ are computed by $f(x)$ and $f^*(x)$. The training data with 2000 $(x_i, y_i)$ pairs is chosen to perform the smooth function approximation with two fine-feature intrusions. MLPs with 1:15:1 architectures are applied.

3). Define a smooth function by
\begin{eqnarray}
F_3(x) = g(x, \frac{1}{5}, \frac{1}{4}, \frac{1}{12}) + g(x, \frac{1}{5}, \frac{3}{4}, \frac{1}{12}) + g(x, \frac{1}{64}, \frac{5}{4}, \frac{1}{12}) \nonumber
\end{eqnarray}

The outliers from the under-sampled points are given by the lower sampling frequency. $x \in X=[0, 1.5]$, the outliers $x^*_i$ are collected by sampling 200 numbers from a uniform distributed grid on $[0, 0.5]$ and 200 numbers from a uniform distributed grid on $[1.0, 1.5]$.  The smooth function is generated by 2000 numbers of points from a uniform distributed grid on $(0.5, 1.0)$. So the oracle function is $f_3(x) = F_3(x), x \in (0.5,1)$ and the outlier generating function is $f_3(x) = F_3(x), x \in [0,0.5]\bigcup[1,1.5]$. The output $y_i$ are computed by $F_3(x)$. The training data with 2400 $(x_i, y_i)$ pairs is the function contaminated by the unevenly under-sampled segments. MLPs with 1:12:1 architecture are initiated to all training sessions.

The sample functions represent three typical sources of outliers (Figure \ref{fig:funcapprox} (a)). The outliers in $F_1(x)$ are in the range of the uncontaminated data but appearing at wrong positions. This sometimes occurs when the labels are normal but maybe corrupted or inaccurate. $F_2(x)$ is affected by outliers with significant large deviations (even exceeding the bounds), indicating the situations where the labels contain some unknown samples. $F_3(x)$ includes seemingly uncontaminated but shifted and under-sampled data. This may happen among heterogeneous observations from several experiments where the samples are highly sparse and biased, and the main function needs to get rid of their influence.
   
In the following experiments, we use both fixed $\lambda$ and adaptive learning methods to update the RO index. $\lambda$ starts at $-10^3$. NAAE is optimized with batch gradient descent. Batch size is fixed at 20. For each of the functions, we generate the test set consisting 2000 samples from the oracle function $f_i(x), (i \in {1,2,3})$ respectively. 

The approximation results of 10 runs are summarized in Table \ref{tab:funcapproxstats}. NAAE is more robust to these three types of outliers with both update strategies. We did not find significant differences between fixed-$\lambda$ and adaptive learning on these experiments.  Training with MSE always achieves lower training error but with a higher test error, which is also known as overfitting. NAAE eliminates the impact of outliers efficiently to achieve better test errors. It is interesting that when approximating the smooth function with fine-features ($F_2(x)$), NAAE achieves better training and test error simultaneously. This perhaps due to both the robustness and optimality of NAAE. Optimizing MSE is unable to get rid of the outliers. The high non-convexity of the compound function also leads to the problem of local optimum. While ignoring the outliers, NAAE seeks the better local optima due to the expansion of the convex region to obtain better performance. The empirical breakdown point of NAAE on these three samples are 0.4\%, 8.4\% and 16.7\%. The training results are shown in Figure \ref{fig:funcapprox}.       

\subsection{Visual Digits Recognition}
We further investigate NAAE in a typical learning problem, the digits recognition tasks on the MNIST dataset. The MNIST dataset \cite{lecun1998gradient} consists of hand written digits 0-9 which are 28x28 in size. There are 60,000 training images and 10,000 testing images in total. We use 10,000 images in the training set for validation to select the hyperparameters and report the performance on the test set. We test our method on this dataset without data augmentation. 

The NAAE function is minimized by batch gradient descent with momentum at 0.9. The learning rate and $l_2$ penalty are fixed at $0.5$  and $0.05$. The penalty weight $a$ are selected in $\{1, 0.1, 0.001\}$ on validation sets respectively. The initial $\lambda$ is fixed at -100. We use the hold-out validation set to select the best model, which is used to make predictions on the test set. All experiments are implemented quite easily in Python and Theano to obtain GPU acceleration \cite{Bastien-Theano-2012}.

\begin{savenotes}
\begin{table}[t]
	\centering
	\begin{tabular}{ll}
		\toprule
		Method\footnote{%(1)\cite{zeiler2013stochastic};(2)\cite{goodfellow2013maxout};
			(1)\cite{mairal2014convolutional};(2)\cite{lee2014deeply};
			(3)\cite{zeiler2013stochastic};(4)\cite{jarrett2009best}} & Error \% \\
		\midrule 
		%    Convolutional DBN \cite{lee2009convolutional} & 0.82 \\
		%    Convolutional NIN + dropout \cite{Lin2014Net} & 0.47 \\
		%ConvNets + Stochastic pooling + dropout$^{(1)}$  & 0.47 \\
		%ConvNets + Maxout + dropout$^{(2)}$ & \textbf{0.45} \\
		Convolutional Kernel Networks$^{(1)}$  & \textbf{0.39} \\
		Deeply Supervised Nets + dropout$^{(2)}$  & \textbf{0.39} \\
		%    Convolutional Kernel Networks + L-BFGS-B \cite{mairal2014convolutional} & \textbf{0.39} \\
		%    Deeply Supervised Nets (DSN-L2SVM) + dropout \cite{lee2014deeply} & \textbf{0.39} \\   
		\midrule
		\midrule
		ConvNets + dropout$^{(3)}$   & 0.55  \\
		large ConvNets, unsup pretraining$^{(4)}$   & 0.53 \\
		ConvNets + NAAE (Ours) & 0.53 \\
		ConvNets + NAAE + dropout (Ours) & \textbf{0.39} \\
		\bottomrule
	\end{tabular}%
	\caption{Test set misclassification rates of the best methods that utilized convolutional networks on the original MNIST dataset using single model.}
	%  Our approach achieves the state-of-the-art result among the algorithms that use standard ConvNets.}
	\label{tab:MNISTresults}%
\end{table}%
\end{savenotes}

On the MNIST dataset we use the same structure of LeNet5 with two convolutional  max-pooling layers but followed by only one fully connected layer and a densely connected softmax layer. The first convolutional layer has 20 feature maps of size $5 \times 5$ and max-pooled by $2 \times 2$ non-overlapping windows. The second convolutional layer has 50 feature maps with the same convolutional and max-pooling size. The fully connected layer has 500 hidden units. An $l_2$ prior was used with the strength $0.05$ in the Softmax layer. Trained by ANRAT, we can obtain a test set error of 0.53\%, which is the best result we are aware of that does not use dropout on the pure ConvNets. With dropout, our method achieve the same performance with the state-of-art at 0.39\% error. We summarize the best published results on the standard MNIST dataset in Table \ref{tab:MNISTresults}.
	
The above results demonstrate both the optimality and robustness of NAAE. The improvements in  MNIST are generally due to better regularization techniques with good optimization strategy. With simple experiment settings and network architectures, NAAE enables the simple ConvNets to optimally learn the models while getting rid of specific outlier samples to enhance the generalization capability. 

\begin{figure}[t]
	\vspace{.in}
	\centering
	\includegraphics[width=3.3 in]{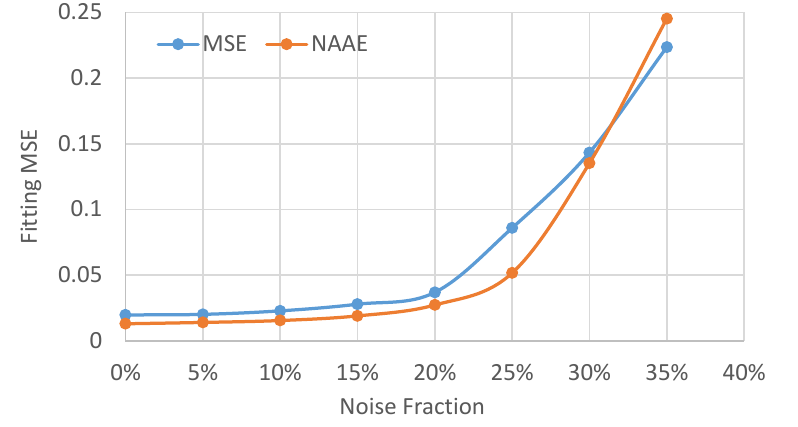}
	\caption{Digit recognition error rates versus percent corrupted labels.} 
	\label{fig:noisemnist}
	\vspace{.3in}
\end{figure}

To better evaluate the robustness of NAAE, we used a fraction of a permutation of the labels to add outliers among the samples but without fixed patterns like \cite{reed2014training}. The learning model is the same as the last experiment on the MNIST dataset. The noise fraction ranges from 0\% to 35\%. Figure \ref{fig:noisemnist} shows that our NAAE with the adaptive training method provides a significant benefit in the case of permuted labels. The consistently lower fitting MSE also indicates its optimality. As the noise level increases up to 30\%, NAAE provides the benefit in this high-noise regime, but is only slightly better than or same with training using MSE overall. When the noise fraction is larger than 30\%, NAAE perform even worse than MSE due to the failure of robustness to identify the outliers, while the convex region still expands to 'optimally' fit the noisy model, thus incurs overfitting.     

\section{Conclusions}
We introduced the Normalized Anomaly-Averting Estimator by pushing the robust-optimal (RO) index $\lambda$ to $-\infty$. It is robust to outliers due to its quasi-minimin functionality. The robustness is realized and controlled by its adaptive RO index without any predefined threshold. Its optimality is guaranteed by the expansion of the convexity region in its Hessian matrix to largely avoid the local optima. We provide both theoretical and empirical support for its robustness and optimality.

In future work, it may be promising to consider updating $\lambda$ between $-\infty$ to $\infty$ to achieve robustness to both small noise and large outliers while preserving optimality, provide the risk and population risk bounds and extend our approach to other areas like aircraft and robotics control. It is also interesting to augment large-scale training for detection (e.g. ILSVRC and speech) with unlabeled and more weakly-labeled images/corpus.

\newpage

\small
\bibliographystyle{apalike}
\bibliography{NAAE}

\end{document}